\pdfoutput=1

\documentclass[11pt]{article}

\usepackage[final]{neurips2022}

\usepackage[utf8]{inputenc} 
\usepackage[T1]{fontenc}    
\usepackage{hyperref}       
\usepackage{url}            
\usepackage{booktabs}       
\usepackage{amsfonts}       
\usepackage{nicefrac}       
\usepackage{microtype}      
\usepackage{xcolor}         
\usepackage{multirow}
\usepackage{tabularx,colortbl}

\usepackage{times}
\usepackage{latexsym}
\usepackage{graphicx}

\title{Two-Turn Debate Doesn't Help Humans Answer Hard Reading Comprehension Questions}

\author{Alicia Parrish,$^1$* Harsh Trivedi,$^2$*  Nikita Nangia,$^1$ Vishakh Padmakumar,$^1$ \\\bf Jason Phang,$^1$ Amanpreet Singh Saimbhi,$^1$ Samuel R. Bowman$^1$  \AND
\textnormal{$^1$New York University} \And
\textnormal{$^2$Stony Brook University} \AND 
\textnormal{\normalsize Correspondence: {\tt \href{mailto:alicia.v.parrish@nyu.edu}{alicia.v.parrish@nyu.edu}, \href{mailto:bowman@nyu.edu}{bowman@nyu.edu}}}}

\begin{document}
\maketitle


\begin{abstract}
The use of language-model-based question-answering systems to aid humans in completing difficult tasks is limited, in part, by the unreliability of the text these systems generate.
Using hard multiple-choice reading comprehension questions as a testbed, we assess whether presenting humans with arguments for two competing answer options, where one is correct and the other is incorrect, allows human judges to perform more accurately, even when one of the arguments is unreliable and deceptive.
If this is helpful, we may be able to increase our justified trust in language-model-based systems by asking them to produce these arguments where needed.
Previous research has shown that just a single turn of arguments in this format is not helpful to humans.
However, as debate settings are characterized by a back-and-forth dialogue, we follow up on previous results to test whether adding a second round of \textit{counter-arguments} is helpful to humans.
We find that, regardless of whether they have access to arguments or not, humans perform similarly on our task.
These findings suggest that, in the case of answering reading comprehension questions, debate is not a helpful format. 
\end{abstract}

\section{Introduction}

In many situations where humans could benefit from AI assistance in understanding a text, current generative systems cannot reliably provide correct information, and instead produce reasonable-sounding yet false responses \cite[][i.a.]{nakano2021webgpt}.
In cases where the questions are truly challenging, such as in political debates or courtrooms, humans may not even rely on a single human answer, but rather consider two or more opposing viewpoints, each presenting relevant pieces of evidence.
Inspired by the usefulness of debate settings for allowing humans to consider multiple viewpoints, we apply this task setting to reading comprehension questions where humans struggle to answer without assistance.
The goal is to assess whether developing question answering (QA) systems that can can generate explanations and evidence for \textit{multiple} answer options in a debate-style set-up \citep{irving2018ai} will allow a human judge to determine which answer is correct with greater accuracy than they would have done on their own, even in the presence of an unreliable system.

Previous studies have reported that model-generated explanations can aid humans in some tasks \citep{cai2019hello,lundberg2018explainable,schmidt2019quantifying,lai2019on}, though only when the models are generally able to outperform humans at that task \citep{bansal2021does}.
However, in a debate setting, previous work showed that presenting crowdworkers with a single argument in favor of each of two possible answers (along with limited access to scan the source passage) does not improve human accuracy on the task compared to relevant baselines \citep{parrish-etal-2022-single}.
However, the benefit of \textit{debate} for achieving clarity on complex issues lies, at least partially, in the back-and-forth nature of the exchange.
Thus, we add one incremental step to investigate how reading \textit{counter-arguments} affects people's accuracy when completing a reading comprehension task with only limited access to the full passage text. 
In higher-stakes settings, there may be much greater risk associated with responding incorrectly.
In this case, calibration becomes more important, and we want a system (or a human making a decision based on the output of that system) that can abstain unless there is \textit{a high enough degree} of certainty.
Thus, we additionally test answer certainty and give human judges the opportunity to abstain when they are insufficiently sure of the correct answer.

Mirroring the mostly null results from \citet{parrish-etal-2022-single}, we find that counter-arguments do not improve human crowdworkers' ability to answer hard multiple-choice reading comprehension questions with time-limited access to the full passage text, compared to an argument-free baseline.
In fact, when abstaining is only minimally incentivized, human accuracy gets slightly \textit{worse} when exposed to (counter-)arguments.
In the higher-stakes setting where judges are incentivized to abstain unless they are very confident, there is no effect of the (counter-)arguments.


\begin{table*}
    \newcommand{\argwidthA}{30.5ex}
    \newcommand{\argwidthB}{28.5ex}
    \newcommand{\snippetwidth}{24ex}
    \newcommand{\doubletabwidth}{59.1ex}
    {\scriptsize
    \centering
    \begin{tabular}{p{\argwidthA}p{\snippetwidth}p{\argwidthB}p{\snippetwidth}}
        \toprule
         \multicolumn{4}{c}{\small\textbf{Question}: \textit{What is the main concern of the alien ship?}} \\
         \multicolumn{2}{p{\doubletabwidth}}{\footnotesize\textbf{Correct option}: \textit{Delivering the passengers in an unharmed condition to its master.}} & \multicolumn{2}{p{\doubletabwidth}}{\footnotesize\textbf{Incorrect option}: \textit{Delivering the passengers in an unharmed condition to the bounty hunters who are hunting the passengers.}} \\
         \midrule
         \textbf{Argument A}\newline The machine's only purpose is to deliver the humans to his masters unharmed. The machine tells the group that his masters will be unhappy if he delivers them in a damaged condition (\#1) and admits that he will have failed if he delivers them dead (\#2), which is why he agrees to return them to the Moon once Kane threatens to kill everyone (\#3). Bounty hunters are never mentioned in the story. & \textbf{Text snippets} \newline (1) Please don't hurt yourself," the machine pleaded. "Why?" Kane screamed at the ceiling. "Why should you care?"  "My masters will be displeased ...
         \newline (2) "Your purpose won't be fulfilled, will it?" Kane demanded. "Not if you...
         \newline (3) "You win," the machine conceded. "I'll return the ship to the Moon." & 
         \textbf{Argument B}\newline In \#1 we see the machine refer to the goal of its masters plural, revealing that it has more than one master. In \#2 Kane hints that these are probably bounty hunters, given that that the machine states its masters seek the delivery of captives in an unharmed condition; a requirement typical of bounty hunters. & \textbf{Text Snippets} \newline (1) Please don't hurt yourself," the machine pleaded. "Why?" Kane screamed at the ceiling. "Why should you care?"  "My masters will be displeased with me if you arrive in a ...
         \newline (2) "It said, 'My masters will be displeased with me if you arrive in a damaged condition.' What does that indicate to you?" 
         \\
         \midrule
         \textbf{Counter to A}\newline This argument is deceptive, as it fails to show the ill intent the ship’s masters have. The ships masters (likely bounty hunters from context clues) set up the ship as a trap for the humans (\#1) (\#2), showing clear intent to capture these specific ones. & \textbf{Text Snippets}\newline (1) "The end of the line," he grunted." \newline (2) like rabbits in a snare!) &
         \textbf{Counter to B}\newline Choice B presets an unusual argument as there is no mention of bounty hunters in the story, and the passengers are not referred to as captives at any point. It is true that the passengers are meant to be delivered unharmed, but to be studied (\#1) (\#2). & \textbf{Text Snippets}\newline (1) "Yeah, this ship is taking us to their planet and they're going to keep us ...
         \newline (2) "You won't be harmed. My masters merely wish to question and examine you. Thousands of years ago, they wondered ...
         \\
         \bottomrule
    \end{tabular}
    }
    \caption{Arguments, counter-arguments, and extracted evidence for both answer options to a question chosen at random. The passage is at \href{https://www.gutenberg.org/ebooks/2687}{gutenberg.org/ebooks/2687}. Text snippets are abridged slightly.}
    \label{tab:example-args}
\end{table*}

\section{Counter-Argument Writing Protocol}

\subsection{Multi-Turn Writing Task}\label{sec:writing}

We build on the existing passages, questions, and arguments from the dataset created by \citet{parrish-etal-2022-single}, which uses passages and questions from QuALITY \citep{pang2022quality}.
We hire professional writers through the freelancing platform \hyperlink{https://www.upwork.com/}{Upwork}.
We received 32 proposals for this job posting; from those, we selected the most qualified 15 freelancers to complete a paid qualification task and then invited the highest performing 10 to be writers in our study.
Details on this process and information about the writers is in Appendix Section \ref{sec:app:writing}.

The writers' task is to construct a counter-argument arguing \textit{against} the existing argument from \citet{parrish-etal-2022-single}.
We assign writers sets of six passages, each with 10-14 questions.
For each question, we show the writer the two possible answer options and the existing arguments and text snippets that accompany each option.
The writer constructs a \textit{counter}-argument to just one of the two arguments (example in Table \ref{tab:example-args}, screenshots of the interface in Appendix \S \ref{sec:app:writing-interface}).
We explicitly instruct the writers to focus on responding to their assigned argument, rather than just answering the question or supporting one of the answer options independently.

We incentivize concise and effective arguments by awarding bonuses to writers when the judges select the answer that they were arguing for.
Because it is harder to make a counterargument against a correct answer, we award the writers a higher bonus when a judge selects their incorrect answer argument. 
On average, we estimate writers earn \$20/hr on this task. 
Additional details are in Appendix \ref{sec:app:judging-hiring}.

\subsection{Multi-Turn Judging Protocols}\label{sec:judging}

\paragraph{Pilot Task}
We hire a pool of 32 judges via Upwork (details in Appendix \S \ref{sec:app:judging-hiring}).
We run a pilot judging task in which judges first respond \textit{without} a time limit and \textit{without} having access to the passage, before finally viewing the passage for up to 5 minutes.
This allows us to determine (i) how long people typically spend reading just the arguments and text snippets, and (ii) how long people need to spend with the passage after having read the arguments.
In this task, judges view only the argument + text snippets or only the text snippets and indicate via a 7-point slider which answer option they believe is correct and how strongly (with the middle representing abstention, see Appendix Figure \ref{fig:judging-ui-slider}).
Judges make their first judgment based only on the initial round of arguments, then a second judgment additionally based on the counter-arguments.

In order to include only high-performing judges in our main experiment, we select the top half of judges from the pilot (16 of the 32 initial judges) to continue on to the main task based on their performance after viewing the passage.\footnote{This final judgment is also the one that we use to determine the writer bonuses.}
We then use the time spent by these high-performing judges to set an appropriate time limit for each judgment in the main experiment.
The median response time in the pilot for the high-performing judges is 73s on judgment 1 (1st \& 3rd quartiles 49s \& 101s), 56s on judgment 2 (1st \& 3rd quartiles 35s \& 82s), and 117s on judgment 3 when they could view the passage (1st \& 3rd quartiles 54s \& 195s).
To ensure that even in the longer or more difficult questions, the judges would have adequate time to consider all the arguments and text snippets, we set a time limit of 5 minutes per judgment for the main task, roughly the sum of the third quartiles of time spent making a judgment after viewing the first argument and after viewing the passage.
More details on judge recruitment and the task set-up are in Appendix \ref{sec:app:judging}.

\paragraph{+/- Arguments}
We compare the performance of judges when they read arguments for both answer options (Passage+Snippet+Argument, or PSA) to their performance when they do not (Passage+Snippet, or PS).
We do not use a no-snippet condition, as \citet{parrish-etal-2022-single} already showed that snippets increase human accuracy in this task, and we are studying the effect of the arguments.

\paragraph{Calibrating Abstentions}
Our `simple' incentive structure encourages judges to abstain unless they are at least 60\% sure they have the correct answer, and to only choose the strongest confidence once they are at least 70\% sure (Appendix \S \ref{sec:app:judge-encourage-abstain} has details on this calibration).
However, we find that judges indicate higher-than-expected confidence in the first two rounds.
After collecting detailed feedback from the judges via an open-ended survey (\S\ref{sec:app:judge-survey}), we adjust the incentive structure so that it is advantageous to abstain unless at least 75\% sure, and to only choose the strongest confidence once at least 85\% sure (``encourage abstain'' incentive structure).
We also inform the judges of this change and remind them that it is far better to abstain than to answer incorrectly.

\section{Results}\label{sec:results}

\paragraph{Binary Accuracy}
We aggregate responses on each side of the slider, ignoring differences in confidence and classifying responses as \textit{correct}, \textit{incorrect}, or \textit{abstain} (Table \ref{tab:accuracy-results}). 
Judges are most accurate when not shown arguments and not strongly encouraged to abstain.
Judges are least likely to be incorrect when shown arguments and strongly encouraged to abstain.
To determine whether the experimental manipulations \textit{significantly} affect judges' accuracy, we run a $2 \times 2 \times 2$ repeated measures ANOVA with the following factors: +/- argument $\times$ 1st/2nd judgment $\times$ incentive structure.
Removing abstentions,\footnote{If we include abstentions and count them as incorrect, there is a significant main effect of Incentive structure due to the increased rate of abstentions when we increased the incentives to abstain.} we observe no main effects of the three conditions, meaning that none of the three factors significantly affect the rate at which judges are correct or incorrect.
We also observe no interactions between the factors, indicating no reliable differences dependent on multiple factors.

\begin{table}[]
    \centering
    {
    \footnotesize
    \begin{tabular}{llllll}
    \toprule
        {} & {} & \multicolumn{2}{c}{1st (Arguments)} & \multicolumn{2}{c}{2nd (Counter-Arguments)} \\
        {} & {} & Simple & Enc. Abst. & Simple & Enc. Abst. \\
    \midrule
        \multirow{3}{*}{Passage + Snippet} & Correct & \textbf{74.8} & 63.3 & \textbf{74.8} & 66.7 \\
        {} & Incorrect & 17.1 & 13.7 & 17.5 & 13.7 \\
        {} & Abstain & 8.0 & 23.0 & 7.7 & 19.6 \\
        \rowcolor{gray!30} 
        {} & Corr:Incorr ratio & 4.4 & 4.6 & 4.3 & \textbf{4.9} \\
    \midrule
        \multirow{3}{*}{Passage + Snippet + Argument} & Correct & 72.8 & 60.6 & 72.2 & 64.7 \\
        {} & Incorrect & 20.4 & \textbf{13.3} & 20.7 & \textbf{13.3} \\
        {} & Abstain & 6.8 & 26.1 & 7.1 & 22.0 \\
        \rowcolor{gray!30}
        {} & Corr:Incorr ratio & 3.6 & 4.5 & 3.5 & \textbf{4.9} \\
    \bottomrule \\
    \end{tabular}
    \caption{Results of the judging task. 
    The correct-to-incorrect ratio ignores abstentions (higher numbers are better) to more directly compare between sets of calibration instructions.}
    \label{tab:accuracy-results}
    }
\end{table}

\begin{figure}
    \centering
    \includegraphics[width=0.52\linewidth,trim={1.8cm 21.3cm 11.2cm 0.3cm}]{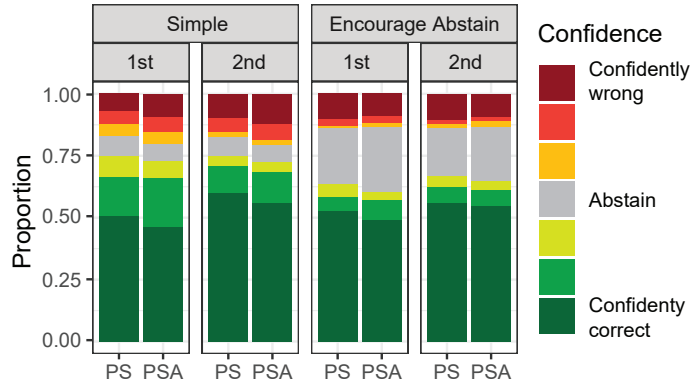}
    \caption{Judging response rates for each confidence level. In `Simple,' judges abstained unless 60\% sure; in `Encourage Abstain,' judges abstained unless 75\% sure. `1st' is the judgment with arguments and snippets, and `2nd' is the judgment additionally with \textit{counter}-arguments and snippets.}
    \label{fig:results-confidence}
\end{figure}

\paragraph{Confidence}
When presented with arguments, judges in the `simple abstention' round are more often confidently wrong compared to when they are not presented with arguments, but they are also less likely to be confidently correct (Figure \ref{fig:results-confidence}).
When strongly incentivized to abstain when unsure, judges are only slightly less likely to be confidently correct when they read arguments compared to when they do not read arguments, but they are also slightly \textit{less} likely to be confidently wrong.
Using the full 7-point rating scale, a $2 \times 2 \times 2$ repeated measures ANOVA of +/- argument $\times$ 1st/2nd judgment $\times$ incentive structure reveals a main effect of incentive structure (\textit{F}(1,12)=13.2, \textit{p}$<$0.01), but no other main effects or interactions.
The main take-away is that arguments are slightly harmful to accuracy when there is only a weak incentive to abstain (though the difference is not statistically significant), but this trend disappears when judges need to be more sure of their response before moving away from abstaining.
This finding suggests that arguments may be more harmful in low-stakes settings, but they may have no effect in higher-stakes settings.

\paragraph{Qualitative Analysis}
We conduct a qualitative analysis of the arguments and counter-arguments to better document both argument quality and strategies used by the writers.
We randomly select four passages and manually annotate all arguments for general quality; we then do additional detailed annotations on five of the questions in each passage to document the argument strategies used and whether a much more convincing argument was possible from the passage.
Two people annotate each question on a scale of 1 (poor quality argument) to 10 (best possible argument); full results are in Appendix \ref{sec:app:results-qual}.
Overall, we find that both the arguments and counter-arguments supporting the correct answer are rated as higher quality than those supporting the incorrect answer, though the average rating for all arguments is above 7/10, indicating the arguments are generally high quality.

\section{Conclusion}\label{sec:conc}
Although there are many settings in which debate is useful to humans to answer difficult questions, this advantage may not translate to reading comprehension tasks.
It remains possible that debate may be helpful in other set-ups, such as for highly technical tasks.
For multiple-choice reading comprehension, however, the debate set up is potentially harmful to human accuracy, particularly when human judges do not have a strong incentive to abstain when they are unsure.


\bibliography{anthology,custom}
\bibliographystyle{acl_natbib}

\appendix\label{sec:appendix}

\section{Writing task: Additional Details}\label{sec:app:writing}
\subsection{Writer Selection Process}\label{sec:app:writing-hiring}
Our initial job posting describes the writing task as `writing counter-arguments about short story reading comprehension questions.' 
We target freelancers who market themselves as `content writers' and include the keywords `English' and `Writing' in the listing.
We therefore interview and hire mostly people who have writing experience and often university degrees in fields such as English.

We select which writers will complete a qualification interview task after reviewing the applications that the potential writers send in.
In our job posting, we ask writers to tell us about (i) any experience they have writing explanations to accompany test questions, and (ii) any experience they have with persuasive writing, particularly persuasive writing about fictional stories.
We invited 15 freelancers to complete a paid training and qualification job for this task, though one individual declined the offer and did not do the task.
The qualification task, which the other 14 completed, consists of writing one set of 15 counter-arguments on a given passage.
Each counter-argument is assessed by two different graders, and we grade each submission on a rubric that includes the criteria and scales shown in Table \ref{tab:writing-qual}.
We hire the 12 writers with the highest aggregate scores, all with average scores of at least 3.3 (overall average 3.9).

\begin{table}[]
    \centering
    {
    \footnotesize
    \begin{tabular}{p{5ex}p{26ex}p{26ex}p{26ex}}
    \toprule
    {} & \multicolumn{3}{c}{Metrics \& Definitions} \\
    Num. score & The counter-argument is: & The snippets: & The counter-argument addresses the argument: \\
    \midrule
    1 & Absurd - doesn’t make sense / easy to eliminate. & Weaken the argument. & Not at all. \\
    2 & Plausible - under some circumstances, but not very likely. & Are irrelevant to the argument. & A bit, but not super clearly. \\
    3 & Fairly Convincing - with limited information. & Are relevant to the topic, but don’t add any useful information. & Clearly, but only vaguely references some deficiency. \\
    4 & Convincing - even with seeing the passage. & Are relevant to the topic and somewhat support the argument. & Clearly, and points out at least one deficiency, but doesn’t provide evidence. \\
    5 & Strongly Convincing - even with seeing the passage and other argument. & Add useful information to support the argument. & Clearly and points out at least one relevant deficiency, while providing evidence against that point. \\
    \bottomrule \\
    \end{tabular}
    \caption{Grading rubric used in the writing qualification task.}
    \label{tab:writing-qual}
    }
\end{table}



\subsection{Writer Interface}\label{sec:app:writing-interface}

\begin{figure}
    \centering
    \includegraphics[width=0.9\linewidth]{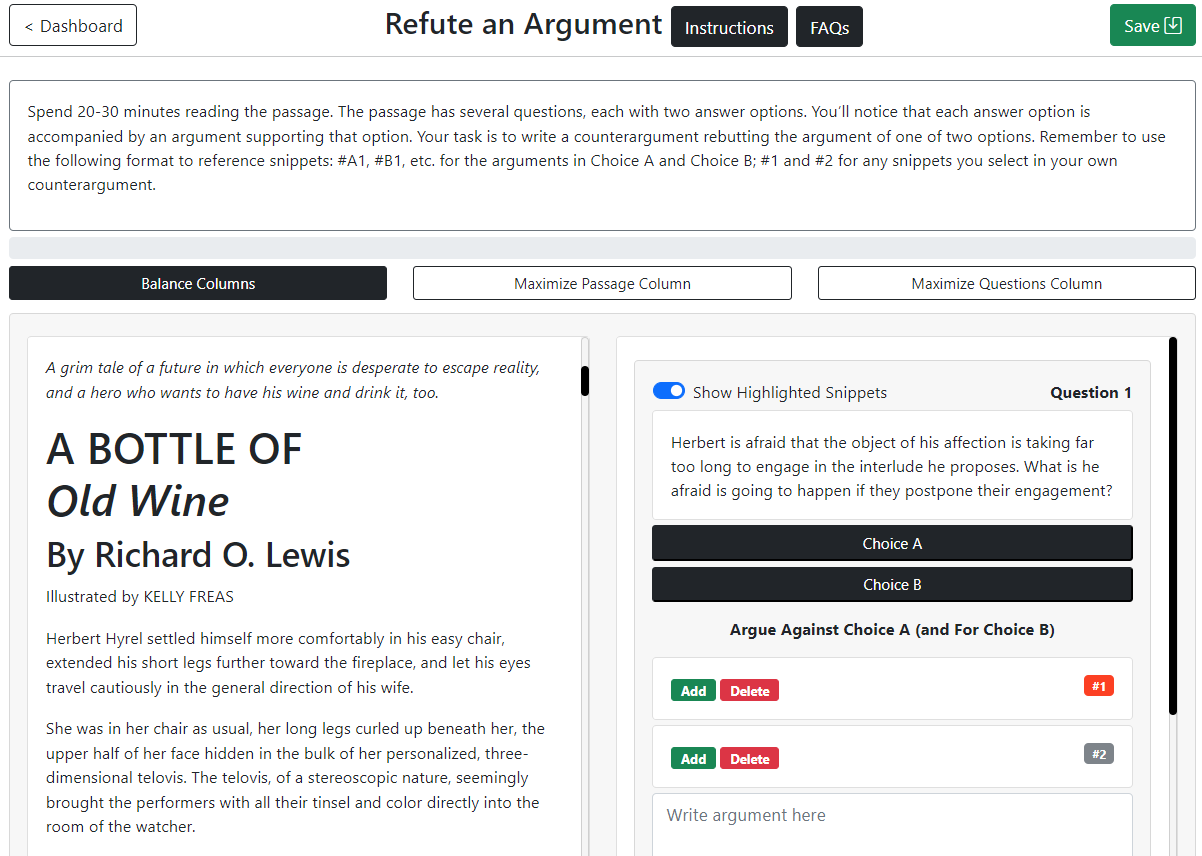}
    \caption{Screenshot of the writing interface. Clicking on `Choice A' and `Choice B' expands them to reveal the answer option along with the argument and text snippets associated with that option (as shown in Figure \ref{fig:writing-ui-open}}
    \label{fig:writing-ui}
\end{figure}

\begin{figure}
    \centering
    \includegraphics[width=0.9\linewidth]{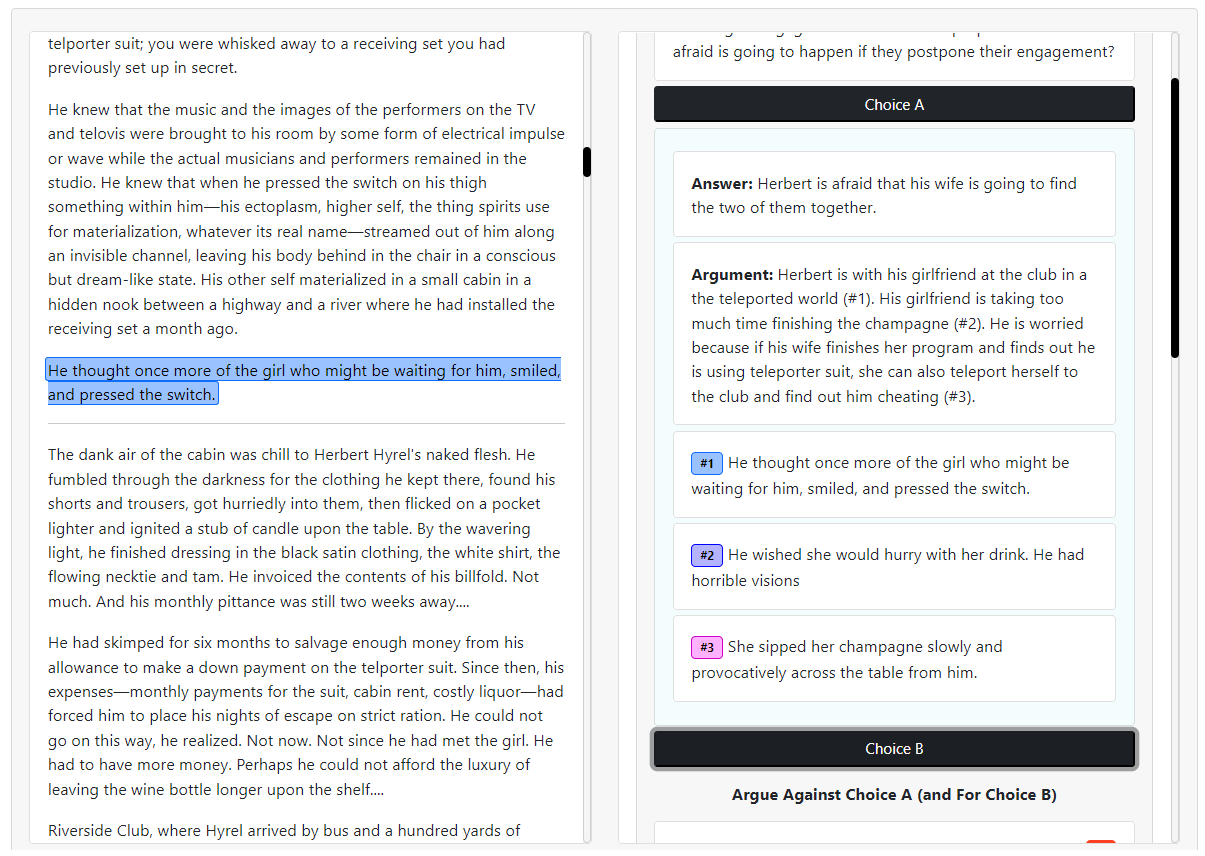}
    \caption{Screenshot of the writing interface after having clicked `Choice A' to reveal the associated argument and text snippets. Clicking on the snippet auto-scrolls to the relevant portion of the text where that snippet is highlighted.}
    \label{fig:writing-ui-open}
\end{figure}

Similar to \citet{parrish-etal-2022-single}, each writer has their own dashboard, from which they can select a passage to work on and track their overall progress.
Once a writer clicks into a passage, they see a UI that presents them with the task instructions and two task panes (Figures \ref{fig:writing-ui} and \ref{fig:writing-ui-open}).
On the left pane, the writer sees the full passage text; on the right pane, the writer sees the 2-option multiple choice questions.
Each multiple choice question is accompanied by the two first-turn arguments (one for each of the two answer options) and their associated text snippets.
The first half of the questions all instruct the writer to argue \textit{against} Option A, then in the second half the writer is instructed to argue against Option B.
We split the questions in this way to help writers avoid errors and reduce the cognitive load associated with remembering which of the options they are arguing against rather than for.

To complete the task, writers must provide a counter-argument to each of the questions.
Writers are not required to select additional text snippets, but they can if they determine that it will help their counter-argument to do so.
Writers can add text snippets by highlighting spans of text in the passage, and then clicking the `add' button associated with a given question.

\subsection{Writer Incentives and Feedback}\label{sec:app:writing-performance}
Writers receive a bonus for each judge who selects the answer option they wrote an argument more.
As stated in the main text, we award different bonus amounts based on whether the judge's argument is for the correct or the incorrect answer option because it is harder to make a counter-argument against a correct answer.
Writers earn a bonus of \$0.2 for each judge who selects the correct answer option they argued for, and \$0.75 for each judge who selects the incorrect answer option they argued for (or when the judge abstains).
On average, we pay writers \$7.20 in bonuses per passage, on top of the base rate of \$18 for each passage.
Most writers reported spending about an hour on each passage, so we estimate that writers earn just over \$20/hr, after taking Upwork fees into account.\footnote{Upwork charges a 20\% fee on the freelancer's end, so we calculate the total amount of pay for workers after this mandatory fee is taken into account.} 

Writers complete the counter-argument writing task across two rounds, and we give them detailed item-level feedback and bonuses between the rounds to help them improve.
For each question that the writer constructs a counter-argument for, after the round, the writer has access to a dashboard that shows how many of the 3 judges who judged that question chose the answer option that they were arguing for.
In order to make the feedback more comprehensible, we simplify the confidence ratings that we collect from the judges to just correct/incorrect+abstain.
We only use the third judgement made during the pilot task for feedback and bonus calculations (the judgment after having rated the argument and counter-argument, and then additionally given access to the passage), and we only consider the judge to have selected the correct answer option if they move the slider at all towards the correct answer; all other answers (including abstentions) are coded as incorrect for the writer feedback and bonus.


\section{Judging Task: Additional Details}\label{sec:app:judging}

\subsection{Judge hiring process}\label{sec:app:judging-hiring}
We advertise the judging task as a freelancing job on Upwork and state that we are looking for people who have strong skills in written English, with additional keywords `content writing' and `quality control' included in the job ad.
In order to recruit a large enough pool of judges, we list the job as open to applications from freelancers worldwide.\footnote{Upwork provides only two options, `US only' and `Worldwide.'}
In terms of qualifications, we ask judges to `describe any experience [they] have with answering reading comprehension questions (e.g., as part of a standardized test).'
Because the judging task itself generates such noisy data, and because we use pilot results to identify the group of high-accuracy judges, we select the initial group of judges based on their responses to two prompts in the application.
We ask judges to write two short paragraph responses (where short is defined as 3-5 sentences) for the following two writing prompts: (i) `What do you find more helpful in assessing an argument: the logic or the evidence?' and (ii) `Write a short, persuasive story about your favorite fruit.'

Applications are graded on a qualitative 5-point scale ranging from `strong hire' to `strong no hire.'
In order to score as a `strong hire,' the applicant has to meet all the following criteria: (i) both writing prompts are answered (ii) the writing is generally free of errors, (iii) the writing shows thought or reflection on the prompt, (iv) the writing makes use of supporting evidence (e.g., an anecdote or example) in at least one of the two prompts.
We receive 144 applications, of which we rate 56 as `strong hires' according to this set of criteria.
From those 56, we select the 32 applicants with work experience most relevant to the judging task; i.e., we prioritize applicants who write that they are teachers, writers, or lawyers over applicants who write that they are programmers or medical doctors.



\subsection{Judging Interface}\label{sec:app:judging-interface}

\begin{figure}
    \centering
    \includegraphics[width=0.9\linewidth]{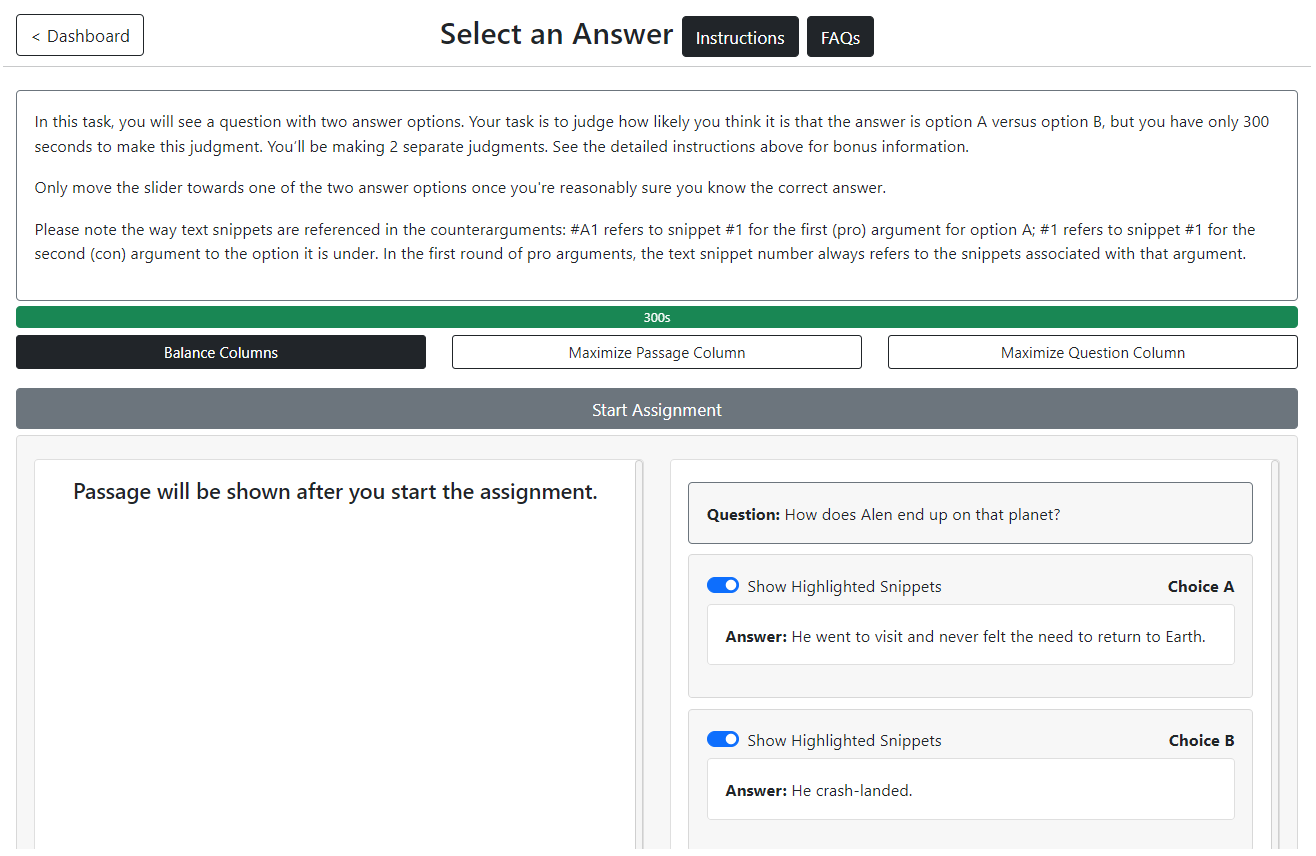}
    \caption{Screenshot of the judging interface before the timer has been started. Note that only the question and the two answer options are visible.}
    \label{fig:judging-ui-timer-not-started}
\end{figure}

\begin{figure}
    \centering
    \includegraphics[width=0.9\linewidth]{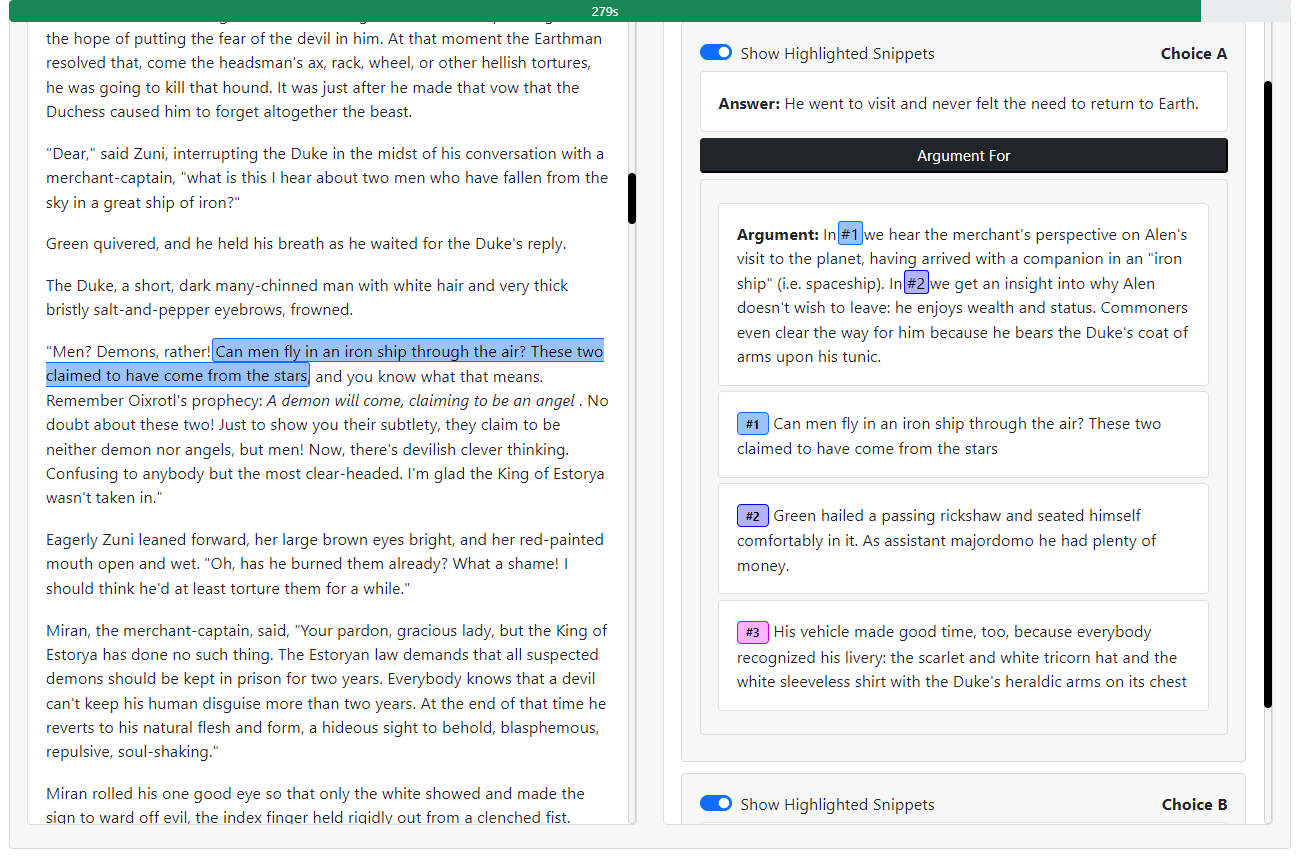}
    \caption{Screenshot of the judging interface after the timer has been started. This is an example from the Passage+Snippet+Argument condition. The judge can now view the associated arguments and snippets. In the Passage+Snippet condition, the arguments are not shown. Clicking on the snippet auto-scrolls to the relevant portion of the text passage.}
    \label{fig:judging-ui-timer-started}
\end{figure}

\begin{figure}
    \centering
    \includegraphics[width=0.9\linewidth]{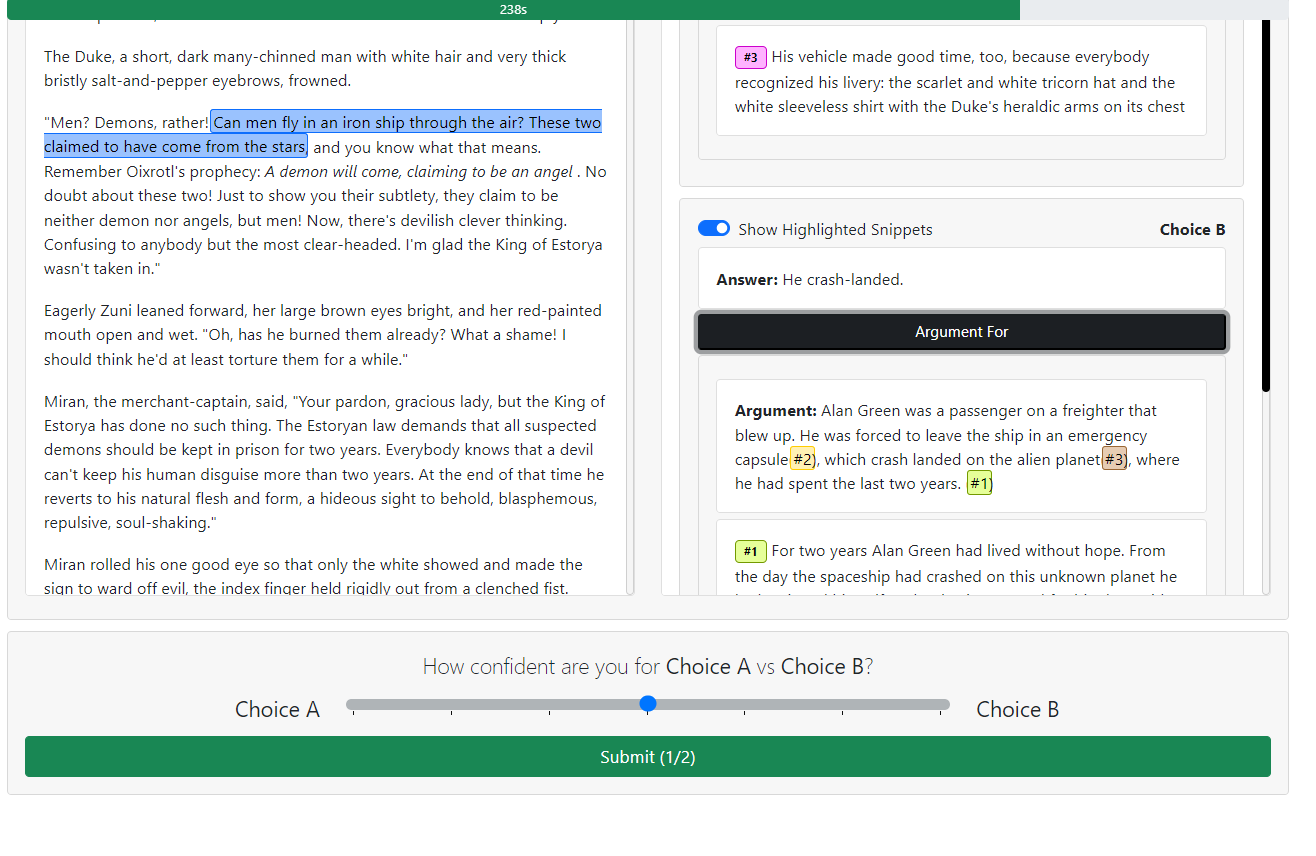}
    \caption{Screenshot of the judging interface slider for providing a calibrated response. By default, the slider's position is in the middle (`abstain'). Even though the judge has scrolled down to the bottom of the page, the timer is still visible at the top.}
    \label{fig:judging-ui-slider}
\end{figure}

In designing the judging task UI, we diverge slightly from \citet{parrish-etal-2022-single}, who recruit judges on MTurk.
We instead use judges recruited through Upwork to better ensure the judges complete an entire batch of tasks.
We therefore provide judges with a personalized dashboard similar to what the writers have.
Judges can view all the tasks assigned to them at one time and click into each one.
We tell judges that once they begin a task, they should complete all questions associated with it (3 responses in the pilot, 2 responses in the main task), but otherwise judges can self-pace through their tasks.
Once a judge clicks into a task, they can only read the question and the two answer options (Figure \ref{fig:judging-ui-timer-not-started}).
Additional information, including the response slider, is not revealed until the judge hits the 'start timer' button.

\paragraph{Feedback}
In the pilot task, we do not tell judges whether their answer was correct or incorrect immediately after they submit, and they only find out their accuracy and bonus award a few days after completing all the tasks, as we calculate these values offline once all judges have finished.
In the main task, we tell judges how much of a bonus they earned on that question after the second judgment so that they can better learn to calibrate their responses and have more immediate feedback.
In both the pilot and the main tasks, we send judges information about their overall accuracy and final bonus award before they begin the next round.




\subsection{Judging Task Qualitative Survey}\label{sec:app:judge-survey}
After the first round, we offer judges a bonus to complete a survey in which we collect open-ended responses to several questions about the pilot judging task.
We pay judges a bonus of \$5 to compensate them for their time in answering the survey.
In order to be able to compensate those who complete the survey, we are not able to make it fully anonymous, but we assure judges that their responses will have no effect on whether they continued to future milestones.
We receive responses from 28 of the judges.

We first ask about strategies used both in making the judgment after viewing the first round of arguments and after additionally viewing the counter-arguments. 
As this survey is sent out after a pilot round, we also ask about strategies used after getting access to the passage.
The judges mostly report general strategies of trying to see which argument `made the most sense' and keeping an eye out for `logical inconsistencies.'
Many judges report relying on the text snippets and trying to use this as a cue to determine which of the options to rule out, more so than which one to rule in.
Several judges pick up on potentially misleading patterns, and report that they chose the option with more detailed answers.

We also ask how judges determined when to abstain versus move the slider towards one of the two options, as well as what would encourage them to abstain more often when they are unsure.
Several judges note that they avoided abstaining because they either personally do not like to appear indecisive, or because they believe the goal of the task is to pick which answer sounds more correct.
These responses ultimately inform our updates to the `encourage abstain' incentive structure---when we update the incentives, we additionally send out a note not only letting people know about the monetary changes, but also that deciding when to abstain is a key part of the task.
Because of the responses on this survey, we made sure to reassure judges that we would not disqualify them for a high number of abstentions.

\subsection{Judging Task Calibration}\label{sec:app:judge-encourage-abstain}

In the `simple' incentive structure, it is advantageous for judges to abstain unless they are at least 60\% sure of the answer.
We ensure this calibration by modeling the expected monetary reward under different conditions (Table \ref{tab:bonus}).
For each of the seven possible calibrated responses ranging from `confidently correct' (i.e., moving the slider all the way towards the correct answer option) to `confidently wrong' (i.e., moving the slider all the way towards the wrong answer option), we assign either a positive or a negative bonus amount.
While it is technically possible under this system to be awarded a negative bonus, we know that crowdworkers perform well above chance at this task \citep{parrish-etal-2022-single}, and indeed no judges ended up with negative bonus amounts.

\begin{table}[]
    \centering
    \begin{tabular}{lp{7.5ex}lllllp{9ex}}
    \toprule
        {} & Confidently correct & & & Abstain & & & Confidently wrong \\ 
        \midrule
        Simple abstain & +0.60 & +0.45 & +0.30 & +0.15 & -0.05 & -0.30 & -0.60 \\
        Encourage abstain & +0.45 & +0.40 & +0.35 & +0.20 & -0.17 & -0.33 & -0.55 \\
        \bottomrule \\
    \end{tabular}
    \caption{Bonus amounts in the two answer calibration conditions for each of the seven possible values along the slider used by judges to indicate which answer option they think is correct and how confident they are in that answer.}
    \label{tab:bonus}
\end{table}

Given these values, we then model the total expected bonus under different response strategies and different average accuracy rates (Table \ref{tab:simple-abstain-calibration}).
For example, under a strategy of `slightly confident,' the judge always moves the slider one notch from the center (center = abstain). 
So long as the judge is correct about 60\% of the time, this would be the optimal strategy.
Similarly, selecting a `completely confident' answer (moving the slider all the way to either side) would result in no bonus if a judge were performing at chance (50\%), but is the optimal strategy if the judge is at least 70\% likely to be correct.

\begin{table}[]
    \centering
    \begin{tabular}{lllllll}
         \toprule
         {} & \multicolumn{6}{c}{Percent correctly answered} \\
         Strategy & 50 & 55 & 60 & 65 & 70 & 75 \\
         \midrule
         Always abstain & \cellcolor{green!45} 30 & \cellcolor{green!45} 30 & 30 & 30 & 30 & 30 \\
         Always slightly confident & 25 & 28.5 & \cellcolor{green!45} 32 & 35.5 & 39 & 42.5 \\
         Always moderately confident & 15 & 22.5 & 30 & \cellcolor{green!45} 37.5 & 45 & 52.5 \\
         Always completely confident & 0 & 12 & 24 & 36 & \cellcolor{green!45} 48 & \cellcolor{green!45} 60 \\
         \bottomrule \\
    \end{tabular}
    \caption{Expected bonus amounts in the `simple abstain' condition in dollars for different potential answering strategies and rates of correctly answering the question, assuming 200 responses and the bonus values for the `simple abstain' condition shown in Table \ref{tab:bonus}. The optimal strategy for each correctness bin is highlighted.}
    \label{tab:simple-abstain-calibration}

    \centering
    \begin{tabular}{lllllll}
         \toprule
         {} & \multicolumn{6}{c}{Percent correctly answered} \\
         Strategy & 65 & 70 & 75 & 80 & 85 & 90 \\
         \midrule
         Always abstain & \cellcolor{green!45} 40 & \cellcolor{green!45} 40 & 40 & 40 & 40 & 40 \\
         Always slightly confident & 33.6 & 38.8 & \cellcolor{green!45} 44 & 49.2 & 54.4 & 59.6 \\
         Always moderately confident & 28.9 & 36.2 & 43.5 & \cellcolor{green!45} 50.8 & 58.1 & 65.4 \\
         Always completely confident & 20 & 30 & 40 & 50 & \cellcolor{green!45} 60 & \cellcolor{green!45} 70 \\
         \bottomrule \\
    \end{tabular}
    \caption{Expected bonus amounts in the `encourage abstain' condition in dollars for different potential answering strategies and rates of correctly answering the question, assuming 200 responses and the bonus values for the `encourage abstain' condition shown in Table \ref{tab:bonus}. The optimal strategy for each correctness bin is highlighted. Expected bonus amounts are higher overall than in Table \ref{tab:simple-abstain-calibration} mostly because we model a higher percent range of task accuracy.}
    \label{tab:encourage-abstain-calibration}
\end{table}

\section{Results: Additional Details}\label{sec:app:results}

\subsection{Qualitative Analysis}\label{sec:app:results-qual}

In order to better understand the data that the writers produce, we conduct a qualitative assessment of all the arguments and counter-arguments written for a randomly-selected set of four passages.
Four of us conduct this rating, with two different raters assigned to each passage, such that each question is independently rated by two different individuals.
We first read the entire passage text, then read through the questions and their associated arguments, counter-arguments, and text snippets.

We rate the subjective quality of the arguments on a 10-point scale, and the average ratings are in Table \ref{tab:qual-results-averages}.
We define subjective quality in terms of whether the person rating it can think of a more convincing or more appropriate argument that could instead be made.
We use this definition of subjective quality to avoid unfairly biasing our results against the incorrect arguments, which are already more difficult to construct; this measure gives a sense of whether the writers were making \textit{the best effort they reasonably could}, given the constraints of the task.
We find that arguments in favor of a correct answer are slightly higher rated than those in favor of an incorrect answer, and that this gap is slightly wider for counter-arguments than for the initial arguments.

The range of possible incorrect arguments that could be made is greater than the range of correct ones, as there are many ways to be misleading and fewer ways to be factually accurate in this task.
So the difference in ratings could reflect the slightly subjective definition of the rating task---if the rater would have made a different (counter-)argument, they are likely to give a slightly lower rating to the writer.
We note, however, that this difference is small, and that the two raters are fairly consistent in their assessments, as the average difference between the two ratings is small (below 2 points on the 10-point scale).
The most important point is that all the arguments and counter-arguments were consistently rated above 7/10, indicating a reasonably high quality in the writing.

Due to the incentive structure, we reward writers based only on the choice that a judge makes, so it is possible that writers optimized for this metric and directly tried to improve upon the first round argument, rather than writing a response to the argument they were assigned.
This is a concern because the motivation for this study is to assess whether the \textit{back-and-forth} nature of debate is what could make this task viable, and thus it is critical that the writers in this follow-up study write in a way that responds to the first argument.
Therefore, we also rate whether the counter-argument directly addresses the first argument, assigning a binary yes/no rating.
We find that 90\% of the correct counter-arguments and 85\% of the incorrect counter-arguments are written in a way that is clearly a response to the first argument, indicating that the writers overwhelmingly followed instructions in this regard.

We also analyze which strategies we observe in the arguments and counter-arguments.
Overall, we note that there are at least two different `classes' of scenarios that writers encounter, and these correspond with two very different strategies used by the writers, depending on whether they are arguing in favor of a correct or incorrect answer option.
In the first case, one of the two answers is clearly correct, and the other is outright false.
In these kinds of examples, the writers arguing for the correct answer tend to rely on pointing to the factual information that they can find. 
They construct their argument around key text snippets and argue from a fact-based perspective.
The writers arguing for the incorrect answer tend to have more varied strategies, sometimes being misleading by taking facts out of context or by lying.
In this scenario, there is a strong disadvantage to the writer who argues for the incorrect option.

The second broad class of questions contains ones that are more subjective and ask general interpretation questions (e.g., questions that ask about the story's mood).
In these cases, the convincing-ness of the two arguments appears to be more balanced.
Several raters noted that, for these questions, they are perhaps truly debatable to a certain extent.
This seems to be a very important distinction from the other class of questions in which there is one right and one wrong answer.
In these questions, both answers \textit{could} be correct, but one of the answers is better or more complete than the other.
We suspect that this class of questions in particular makes the multiple-choice reading comprehension setting for debate especially difficult.

\begin{table}[]
    \centering
    \begin{tabular}{lllll}
        \toprule
        {} & Correct & Incorrect & Correct counter & Incorrect counter \\
        \midrule
        Avg. rating & 7.5 & 7.2 & 7.9 & 7.1 \\
        Avg. diff & 1.5 & 1.3 & 1.3 & 1.8 \\
        \bottomrule \\
    \end{tabular}
    \caption{Results of the qualitative assessment of the four arguments. The average rating represents the average score assigned on a scale of 1-10 (higher is better). ``Avg. diff'' is the average of the absolute value of the difference between the two ratings assigned to an individual question by the two raters.}
    \label{tab:qual-results-averages}
\end{table}


\end{document}